\documentclass[oneside]{article}
\usepackage{graphicx}
\usepackage{amsmath}
\usepackage{amsthm}
\usepackage{amssymb}
\usepackage{enumerate}
\usepackage{harvard}

\usepackage{setspace}

\DeclareMathOperator*{\argmin}{arg\,min}

\newcommand{\tamanio}{\textwidth}

\theoremstyle{definition}

\date{}

\begin{document}

\title{Exploiting Heavy Tails in Training Times of Multilayer Perceptrons:  A
Case Study with the UCI Thyroid Disease Database}

\author{Manuel Cebri\'an\thanks{Manuel Cebri\'an is with the Department of Computer Science, Brown University, Box 1910, Providence 02912, USA (email: mcebrian@cs.brown.edu).} \ and Iv\'an Cantador\thanks{Iv\'an Cantador is with the Department of Computer Engineering, Universidad Aut\'onoma de Madrid,
28049 Madrid, Spain (email:
\{manuel.cebrian, ivan.cantador\}@uam.es).}
}

\maketitle

\begin{abstract}
\label{resumen}
%
The random initialization of weights of a multilayer perceptron
makes it possible to model its training process as a Las Vegas
algorithm, i.e. a randomized algorithm which stops when some
required training error is obtained, and whose execution time is a
random variable. This modeling is used to perform a case study on
a well-known pattern recognition benchmark: the UCI Thyroid
Disease Database. Empirical evidence is presented of the training
time probability distribution exhibiting a heavy tail behavior,
meaning a big probability mass of long executions. This fact is
exploited to reduce the training time cost by applying two simple
restart strategies. The first assumes full knowledge of the
distribution yielding a 40\% cut down in expected time with
respect to the training without restarts. The second, assumes null
knowledge, yielding a reduction ranging from $9\%$ to $23\%$.
\end{abstract}

\paragraph{Keywords:} Stochastic Modeling, Multilayer Perceptron, Heavy
Tail Distribution, Restart Strategy, UCI Thyroid Disease Database.

\section{Introduction}
\label{intro}

The training time of a Multilayer Perceptron (MLP), understood as the time
needed to obtain some required training error, is a random
variable which depends on the random initialization of the MLP weights.

These weights are commonly initialized according to a given
probability distribution, having this choice a significant impact
on the training time distribution
\citeaffixed{Delashmit02,Duch97,LeCun98}{see}. To address this
problem, some weight initialization methods have been proposed
\citeaffixed{Duch97,Weymaere94}{e.g.}. They attempt to reduce
 the training time by applying different probability distributions on the initial weights of the MLP based on
knowledge about the training set.

In this correspondence, a simpler and more general approach which
does not make use of the mentioned information is presented. To do
this, we model the learning process of a MLP as a \emph{las Vegas}
algorithm \cite{Luby93}, i.e. a randomized algorithm which meets
three conditions: (i) it stops when some pre-defined training
error $\delta$ is obtained, (ii) its only measurable observation
is the training time, and (iii) it only has either full or null
knowledge about the training time probability distribution.

Using this modeling, we perform a case study with the UCI Thyroid
Disease database\footnote{The UCI Repository of Machine Learning
Databases, available online at
\mbox{http://www.ics.uci.edu/\textasciitilde
mlearn/MLRepository.html} }, revealing that the time distribution
for learning this pattern recognition benchmark belongs to the
\emph{heavy tail} distribution family. This type of distributions
is regarded as non-standard for its big probability mass of
arbitrary long values.

%

We make use of formal and experimental results which prove that the expected
execution time of a random algorithm with such underlying distribution can be
reduced by using \emph{restart strategies} \cite{Gomes03}. This work adapts
these strategies to the  MLP context: the MLP is trained
during a number of
epochs $t_{1}$. If the required training error $\delta$ is achieved before
$t_{1}$, then the execution finishes. Otherwise, we initialize
again the weights in a randomized way, and re-train the
MLP during $t_{2}$ epochs. The process is iteratively repeated until
the training error $\delta$ is reached, being $t_i$ the restart threshold
(in epochs) after $i-1$ restarts have been performed.

Two different strategies are applied for the determination of optimal restarting
times. The first assumes full knowledge of the distribution
yielding a 40\% cut down in expected time with respect to the training without
restarts. The second assumes null knowledge, yielding a reduction ranging from
 $9\%$ to $23\%$.

The rest of the paper is organized as follows. Section
\ref{thyroid} presents the Thyroid Disease database and provides
evidence of heavy tail behavior when a MLP is trained on it.
Section \ref{restarts} tests the condition to be satisfied by the
probability distribution to profit from restart strategies,
providing an empirical evaluation of two strategies on the
particular case study. Finally, some conclusions and future
research lines are given in section \ref{conclusions}.

\section{A case study: the UCI Thyroid Disease Database}
\label{thyroid}

To motivate the use of restarts in MLP learning, we firstly
present the existence of a high variability in its training time, indicative of
an underlying heavy tail behavior. The evaluation was performed using the UCI
Thyroid Disease database, as a case study.

Table \ref{tablaTiroides} shows the expectations, deviations (and
its ratio) of the numbers of epochs $T$ spent in building a single hidden layer MLP with $n=1,\ldots,8$ units. The MLP was trained
using the well-known Back-Propagation technique with a target
training error $\delta=0.02$. The results shown were computed
using $10$-fold cross validation.

\begin{table}[t]
\centering
\begin{tabular}{|c|c|c|c|c|c|c|c|c|} \hline
  $n$ & 1 & 2 & 3 & 4 & 5 & 6 & 7 & 8 \\ \hline
  $E[T]$      & 8551.7 & 5516.8 & 888.5 & 2339.7 & 1680.2 & 587.6 & 482.4 &
490.5 \\ \hline
  $\sigma[T]$ & 2547.5 & 3885.6 & 1565.5 & 2848.8 & 1355.6 & 55.1 & 296.9 &
464.1 \\ \hline
$\sigma[T]/E[T]$ & 30\% & 70\% & 156\% &  106\% & 79\% & 10\% & 60\% &
95\%\\ \hline
\end{tabular}
\caption{Expectation, deviation (and its ratio) of the number of epochs $T$
spent in the building of a MLP with $n$ hidden units and training
error $\delta=0.02$. The training algorithm was run $1,000$ times for each 
number of hidden units.} \label{tablaTiroides}
\end{table}

The obtained deviations are very large respect to the
expectations for most of the architectures. For the rest
of the experiments, we shall use a MLP with $n=3$ hidden units,
which has the highest relative variability. This will serve as a
proof of concept, although the same behavior is observed in MLPs
with other number of hidden units.

In the following, we give visual evidence that $T$ is heavy
tailed, i.e. that the probability of the training time $T$ being
greater than some number of epochs $t$ has polynomial decay,
\emph{viz.} $P[T>t] \sim C.t^{-\alpha}$, where $\alpha \in (0,2)$,
$C$ is some constant, and $t>0$.

\begin{figure}[t]
\centering \includegraphics[width=\tamanio]{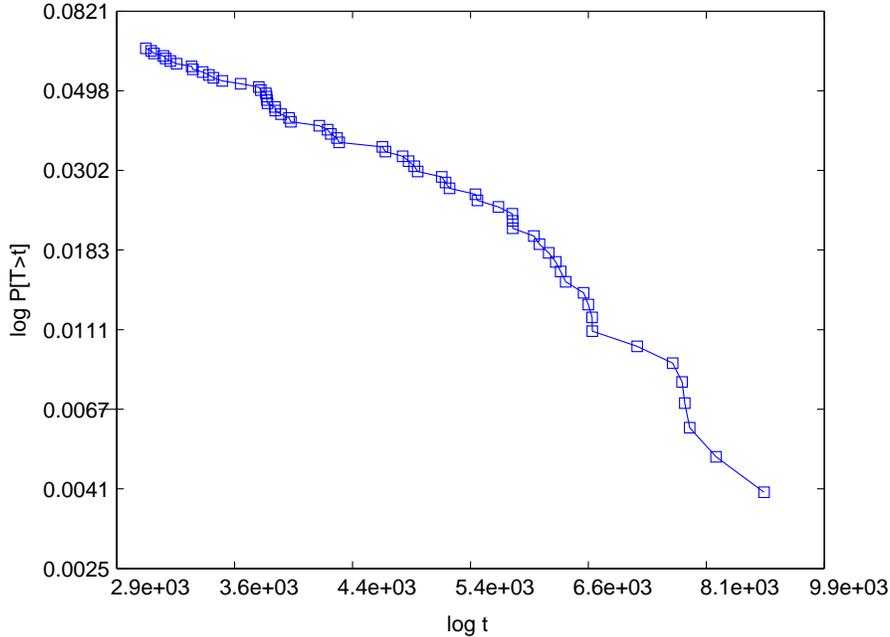}
\caption{A log-log plot of $P[T>t]$ as a function of $t$ (in epochs).}
\label{survivor}
\end{figure}

Figure \ref{survivor} presents a log-log plot of $P[T>t]$ for the
$10\%$ largest values  ($t>3,000$). The plot confirms the polynomial decay by
displaying a straight line with slope $-\alpha$. This is because, for
sufficiently large $t$, $\log P[T>t] = -\alpha \log C.t \Rightarrow \log P[T>t]
/ \log C.t \approx -\alpha$.

Finally, we verify that $\alpha$ belong to the $(0,2)$ interval by
computing the \possessivecite{Hill75}  estimator:

\begin{equation*}
\hat{\alpha}_{r}=\left (r^{-1}\sum_{j=1}^{r}\ln T_{m,m-j+1}-\ln
T_{m,m-r} \right),
\end{equation*}

\noindent where $T_{m,1} \leq T_{m,2} \leq \ldots \leq T_{m,m}$ are the
$m$ ordered training completion times, and $r<m$ is a cutoff that allows to
observe only the highest values (the tail). We use the typical cutoff
$r=0.1m$ and obtain
$\hat{\alpha}_{r}=1.942$, which is consistent with our hypothesis.

This polynomial decay, which yields a big probability mass for
long executions, is due to the fact that certain initial weights
entail a convergence to local minima of the target function,
requiring very long (even infinite) training periods, while others
yield a convergence to global minima in a few epochs.

\section{Restart strategies}
\label{restarts}

A las Vegas algorithm may profit from restarting if, at some point
of the execution $\tau$, the expected completion time conditioned to the
already employed execution time ($E[T-\tau|T>\tau]$) is larger than the
(unconditioned) expected completion time ($E[T]$), i.e. if $\exists \tau, \ \
E[T]<E[T-\tau|T>\tau]$ \citeaffixed{Moorsel04}{see}.

\begin{figure}[t]
\centering \includegraphics[width=\tamanio]{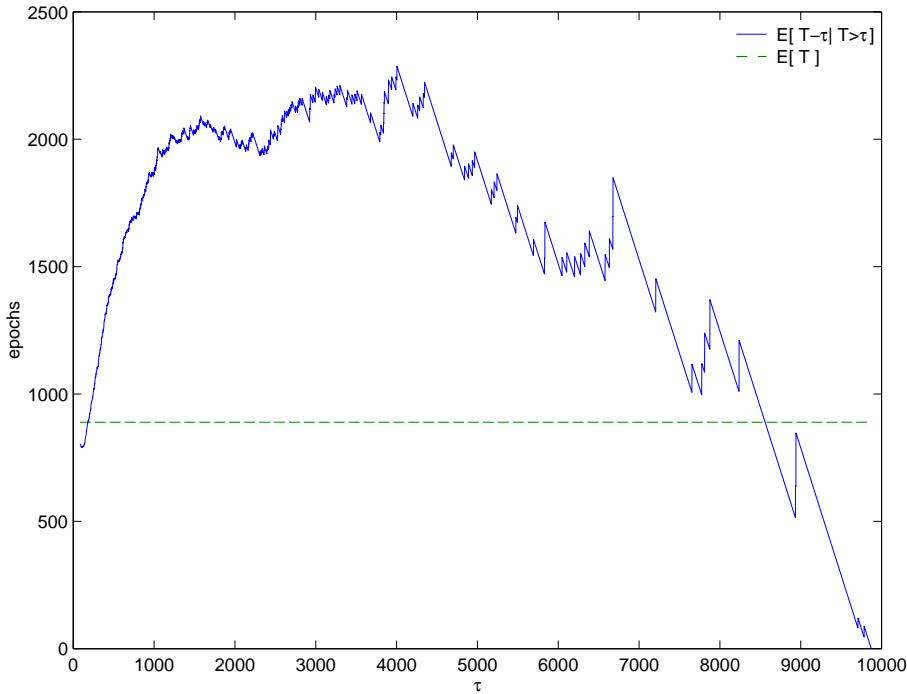}
\caption{$E[T-\tau|T>\tau]$ as a function of $\tau$, $E[T]$ serves as the
baseline.}
\label{condicion}
\end{figure}

Figure \ref{condicion} shows that the majority of $\tau$ values
met the condition for the MLP to profit of restart strategies.

\subsection{Restart strategies when the distribution is known}
\label{known}

\citeasnoun{Luby93} prove the existence of an optimal restart
strategy for a Las Vegas algorithm which minimizes the expected
running time when the execution time distribution $q(t)=\Pr(T<t)$
is assumed known.


This optimal strategy is a fixed restart threshold for al iteration of the form $t_i=t^* \ \forall i$,
where
\begin{equation}
t^{*} = \argmin_t E[S_t]= \argmin_t \frac{1}{q(t)} \left( t-\sum_{t'<t}q(t')
\right)
\end{equation}
and $S_{t}$ is the restart strategy where $t_i=t \ \forall i$ for some $t$. We assume
some discretization of the time, so that expressions like $t'<t$ make sense.

\begin{figure}[t]
\centering
\includegraphics[width=\tamanio]{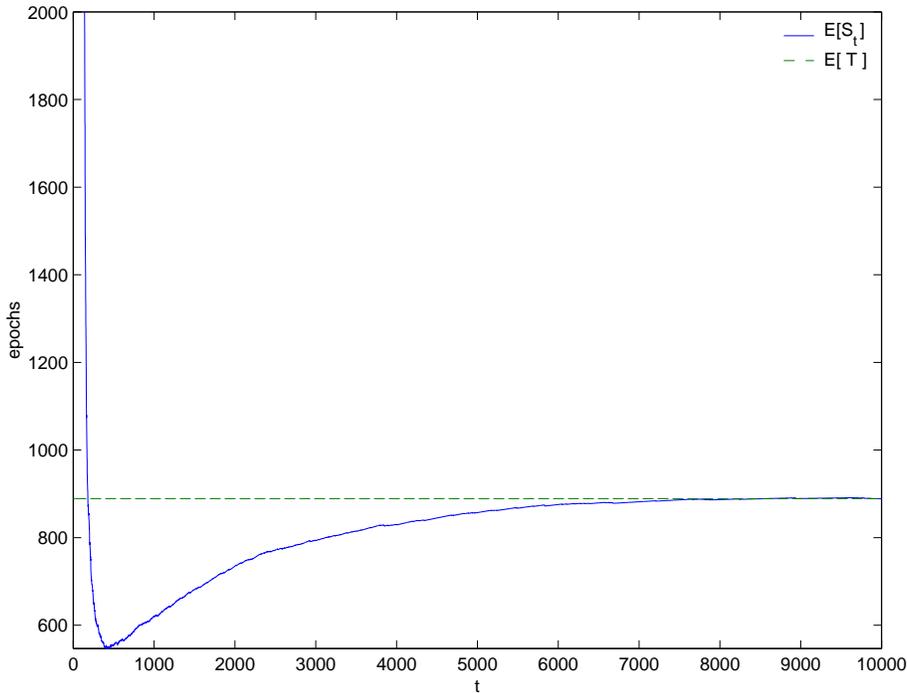}
\caption{Expected training time using the strategy $S_t$ with $t
\in \left[100,\ 10,000 \right]$, $E[T]$ servers as the baseline.}
\label{optima}
\end{figure}

Simple calculations yield $t^{*}=418$, with an optimal expected
time $E[S_t^{*}]=546.876$. This provides a $40\%$ cut down in
expected time with respect to the training without restarts (see
Table \ref{tablaTiroides}). Figure \ref{optima} displays the
expected time for strategies of the form $S_t$ with $t \in \left
[100,\ 10,000 \right]$. As it can be seen, many non-optimal $t$
choices  provide a time reduction as well.



\subsection{Restart strategy when the time distribution is unknown}
\label{unknown}

In some scenarios it is not possible to assume full knowledge of the
distribution, e.g. if the MLP is to be trained a single time. In this
subsection we assume null knowledge.

Again \citeasnoun{Luby93} prove the existence of  an optimal strategy for
this assumption, and \citeasnoun{walsh99search} derives a simpler variant of the
former which is commonly used in practical applications. The Walsh strategy
$S_W$ is defined as
$t_i=\gamma^{i-1},\ \gamma > 1$. This strategy benefits of a
high probability of success when $t_i=\gamma^{i-1}$ is near to $t^{*}$.
Increasing $t_i$ geometrically makes it sure to reach $t^{*}$ in a few
generations, expecting to reach error $\delta$ within few restarts after the
value of $t_{i}$ surpasses the optimal.

\begin{figure}[t]
\centering \includegraphics[width=\tamanio]{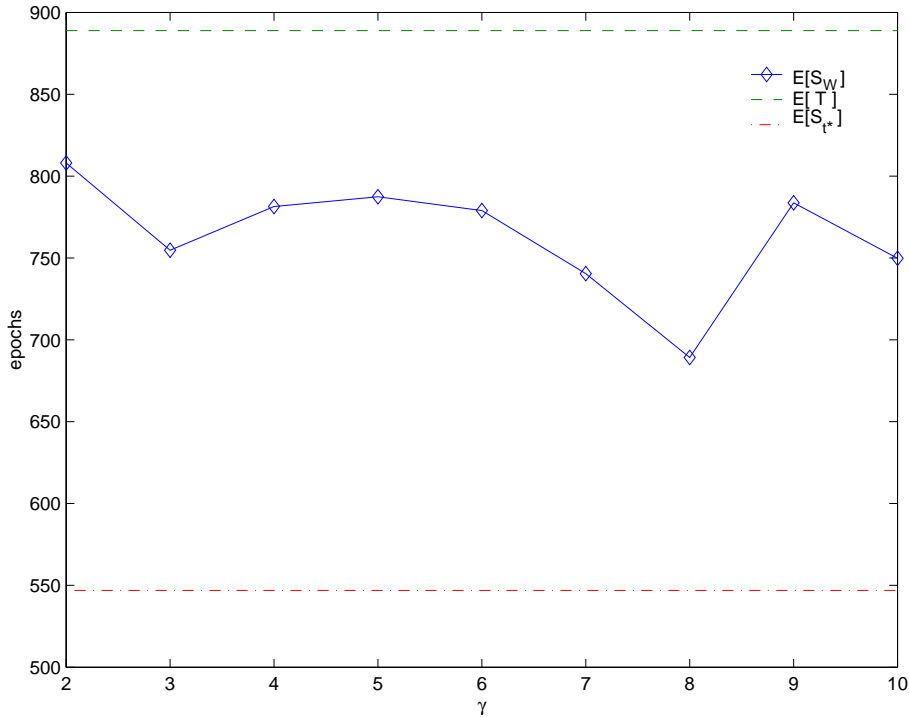}
\caption{Expected training time using the Walsh strategy $E[S_W]$ for
$\gamma=1,2,\ldots,10$,  $E[S_t^*]$ and $E[T]$ serve as baselines.}
\label{gamma}
\end{figure}

Figure \ref{gamma} displays the expected values of
$S_W$ using several standard $\gamma$ values $\gamma=2,3,\ldots,10$. Training is
speeded with all choices, with improvements ranging from $9\%$ ($\gamma =2$) to
$23\%$
($\gamma=8$). The expected times were computed running $1,000$ times the
training algorithm for each $\gamma$.

\section{Conclusions and future work}
\label{conclusions}

In this work, MLP training algorithm is modeled as a Las Vegas
algorithm, performing a case study on the UCI Thyroid Disease
Database. We give visual and numerical evidence that the probability
distribution of the training time belongs to the heavy tail
family, meaning a polynomial probability decay for long
executions. This property is exploited to reduce the training time
cost by two simple strategies. The first assumes full knowledge of
the distribution yielding a 40\% cut down in expected time with
respect to the training without restarts. The second, assumes null
knowledge, yielding a reduction ranging from $9\%$ to $23\%$.

As a future research, we plan to determine whether further
improvements can be obtained by relaxing las Vegas
algorithms assumptions (ii) and (iii) (see section \ref{intro}). This could make
it possible to incorporate dynamic restart strategies
\citeaffixed{Kautz02}{see} capable of exploiting epoch-by-epoch
information about the training time distribution, using various
algorithm behavior measurements besides the execution time.

\section{Acknowledgements}

This work was partially supported by grant TSI2005-08255-C07-06 of the
Spanish Ministry of Education and Science. We would also like to thank Ignacio
Garc\'ia for his useful suggestions and comments on this manuscript.

\bibliographystyle{harvard}
\bibliography{main}

\end{document}